\documentclass[letterpaper]{article} 
\usepackage{aaai25}  
\usepackage{times}  
\usepackage{helvet}  
\usepackage{courier}  
\usepackage[hyphens]{url}  
\usepackage{graphicx} 
\usepackage{natbib}  
\usepackage{caption} 
\usepackage{algorithm}
\usepackage{algorithmic}
\usepackage{amsmath}  
\usepackage{amssymb}  
\usepackage{color}
\usepackage{multirow}
\usepackage[table]{xcolor}

\usepackage{xcolor}         
\usepackage{subcaption}
\usepackage{appendix}

\definecolor{mygreen}{HTML}{39b54a}  
\definecolor{myred}{HTML}{A10035}
\definecolor{myyellow}{HTML}{F8E924}
\definecolor{ggray}{RGB}{127,127,127}
\definecolor{ggreen}{HTML}{3EC70B}
\definecolor{mygray1}{gray}{.5}
\definecolor{mygray}{gray}{.9}
\definecolor{aliceblue}{rgb}{0.94, 0.97, 1.0}

\usepackage{arydshln}

\usepackage{tabulary}
\newcolumntype{I}{!{\vrule width 1pt}}

\newcolumntype{x}[1]{>{\centering\arraybackslash}p{#1pt}}
\newcolumntype{y}[1]{>{\raggedright\arraybackslash}p{#1pt}}
\newcolumntype{z}[1]{>{\raggedleft\arraybackslash}p{#1pt}}
\newlength\savewidth

\usepackage{pifont}
\newcommand{\cmark}{\textcolor{purple}{\ding{52}}}%
\newcommand{\xmark}{\textcolor{blue}{\ding{56}}}%

\definecolor{voc_cow}{HTML}{0C1E7F}
\definecolor{voc_horse}{HTML}{FB2576}
\definecolor{violet}{HTML}{BB1AEF}

\usepackage{tabulary}
\newcolumntype{I}{!{\vrule width 1pt}}
\newcolumntype{x}[1]{>{\centering\arraybackslash}p{#1pt}}
\newcolumntype{y}[1]{>{\raggedright\arraybackslash}p{#1pt}}
\newcolumntype{z}[1]{>{\raggedleft\arraybackslash}p{#1pt}}

\makeatletter
\newcommand{\thickhline}{%
	\noalign {\ifnum 0=`}\fi \hrule height 1pt
	\futurelet \reserved@a \@xhline
}
\makeatother

\newcommand{\increase}[1]{
	{\fontsize{7pt}{0.5em}\selectfont\color{purple}{$\uparrow$~{#1}}}
}

\newcommand{\deincrease}[1]{
	{\fontsize{7pt}{0.5em}\selectfont\color{ggreen}{$\downarrow$~{#1}}}
}

\usepackage{newfloat}
\usepackage{listings}
\DeclareCaptionStyle{ruled}{labelfont=normalfont,labelsep=colon,strut=off} 
\floatstyle{ruled}
\newfloat{listing}{tb}{lst}{}
\floatname{listing}{Listing}
\title{Multi-task Visual Grounding with Coarse-to-Fine Consistency Constraints}
\author {
   Ming Dai\textsuperscript{\rm 1},
   Jian Li\textsuperscript{\rm 2},
   Jiedong Zhuang\textsuperscript{\rm 3},
   Xian Zhang\textsuperscript{\rm 1},
   Wankou Yang\textsuperscript{\rm 1,4 *}
}
\affiliations {
   \textsuperscript{\rm 1}School of Automation, Southeast University, China\\
   \textsuperscript{\rm 2}Youtu Lab, Tencent, China\\
   \textsuperscript{\rm 3}College of Information Science and Electronic Engineering, Zhejiang University, China\\
   \textsuperscript{\rm 4}Advanced Ocean Institute of Southeast Univerisity, Nantong, China\\
   mingdai@seu.edu.cn, swordli@tencent.com, zhuangjiedong@zju.edu.cn, 230218200@seu.edu.cn, wkyang@seu.edu.cn
}

\usepackage{bibentry}

\begin{document}

\maketitle

\begin{abstract}
	Multi-task visual grounding involves the simultaneous execution of localization and segmentation in images based on textual expressions.
	The majority of advanced methods predominantly focus on transformer-based multimodal fusion, aiming to extract robust multimodal representations.
	However, ambiguity between referring expression comprehension (REC) and referring image segmentation (RIS) is error-prone, leading to inconsistencies between multi-task predictions. Besides, insufficient multimodal understanding directly contributes to biased target perception.
	To overcome these challenges, we propose a Coarse-to-fine Consistency Constraints Visual Grounding architecture ($\text{C}^3\text{VG}$), which integrates implicit and explicit modeling approaches within a two-stage framework. 
	Initially, query and pixel decoders are employed to generate preliminary detection and segmentation outputs, a process referred to as the Rough Semantic Perception (RSP) stage. These coarse predictions are subsequently refined through the proposed Mask-guided Interaction Module (MIM) and a novel explicit bidirectional consistency constraint loss to ensure consistent representations across tasks, which we term the Refined Consistency Interaction (RCI) stage.
	Furthermore, to address the challenge of insufficient multimodal understanding, we leverage pre-trained models based on visual-linguistic fusion representations.
	Empirical evaluations on the RefCOCO, RefCOCO+, and RefCOCOg datasets demonstrate the efficacy and soundness of $\text{C}^3\text{VG}$, which significantly outperforms state-of-the-art REC and RIS methods by a substantial margin. 
	Code and model will be available at \url{https://github.com/Dmmm1997/C3VG}.
\end{abstract}

%

\section{Introduction}
Visual grounding is a critical task within the vision-language domain, aimed at establishing a fine-grained correspondence between images and text by grounding a given referring expression within an image~\cite{glip}.
This task is typically divided into two sub-tasks based on the grounding approach: referring expression comprehension (REC)~\cite{mattnet, mdetr} and referring image segmentation (RIS)~\cite{restr, cgformer}.
Traditionally, REC and RIS have been treated as separate tasks with distinct technological pathways, necessitating complex, task-specific designs. However, REC and RIS exhibit significant similarities and offer complementary strengths, making their unification both logical and advantageous.
Recently, multi-task visual grounding has gained prominence as it eliminates the need for task-specific network designs and enables the leveraging of data across both tasks to mutually enhance performance.
MCN~\cite{mcn} was the first approach to jointly train the REC and RIS tasks, employing a learnable method to establish consistency in attention maps.
Recent research has primarily focused on enhancing the interaction across different modalities~\cite{reftr,vg-law} and exploring auto-regressive approaches to achieve both detection and segmentation~\cite{seqtr, pvd, polyformer}. In this paper, we address two overlooked issues: \textbf{1)} \textit{How to effectively leverage the complementarity of multi-task predictions to mitigate inconsistencies in results.} \textbf{2)} \textit{How to overcome the challenge of insufficient multimodal understanding to enhance perception in complex image-text scenarios.}

 \begin{figure}[t]
 	\centering
 	\includegraphics[width=1.0\linewidth]{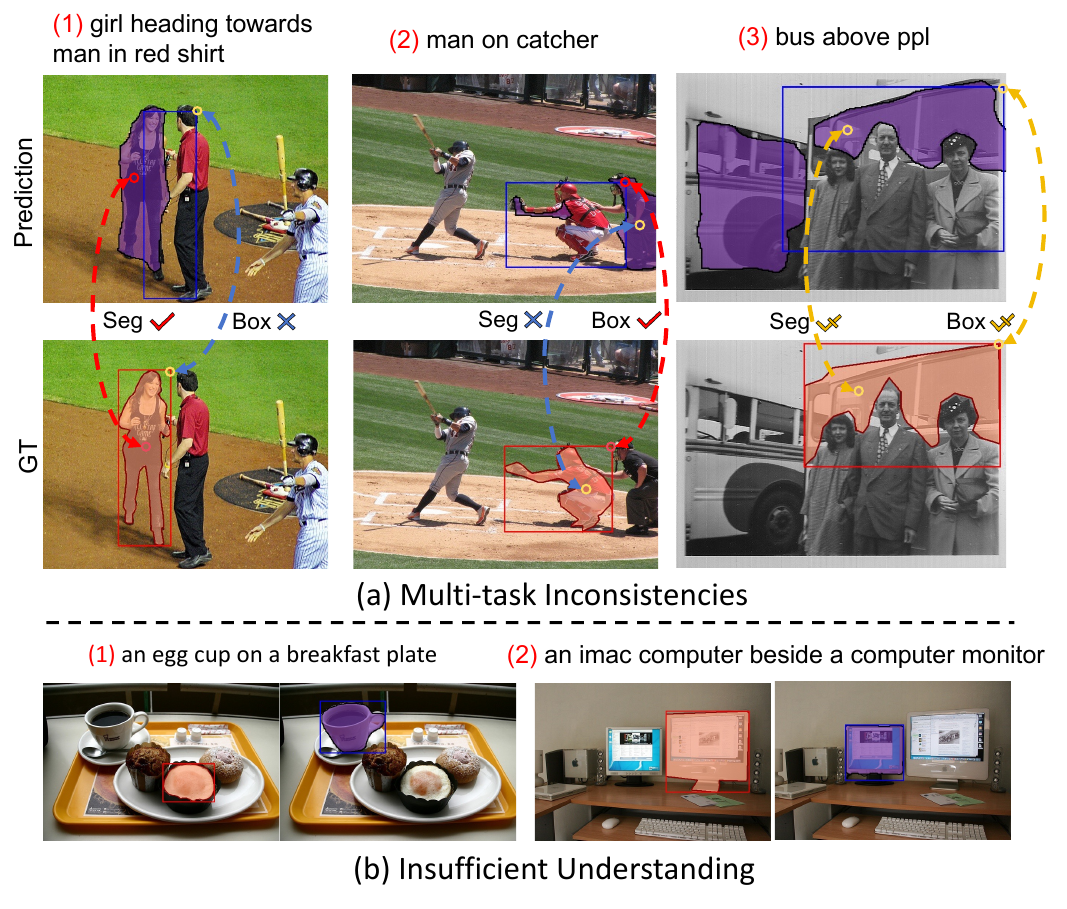} 
 	\caption{
 		(a) Three examples of inconsistent results between multi-task outputs. (b) Two examples of failure in identifying targets due to insufficient multi-modal understanding.
 	}
 	\label{fig:motivation}
 \end{figure}
 
 \begin{figure*}[t]
 	\centering
 	\includegraphics[width=1.0\linewidth]{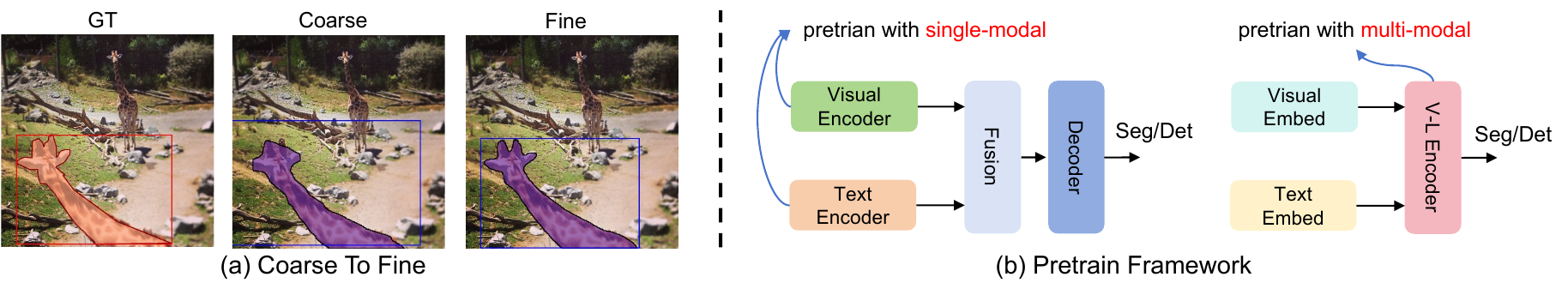} 
 	\caption{
		(a) Examples of the intermediate process in the proposed coarse-to-fine consistency constraint framework.
 		(b) Two pretraining architectures: the left diagram illustrates separate encodings for image and text modalities followed by fusion, using single-modal pretraining; the right diagram shows a fused encoding architecture with multimodal pretraining.
 	}
 	\label{fig:motivation2}
 \end{figure*}
 
\textbf{Inconsistent predictions between multi-task} primarily arise due to the lack of effective constraints linking different tasks. This issue can be exemplified by three scenarios depicted in Fig.~\ref{fig:motivation}(a): (1) accurate segmentation but erroneous detection, (2) inaccurate segmentation but correct detection, and (3) both segmentation and detection being incorrect yet providing complementary information.
The traditional REC is a one-to-one detection task. When uncertainties arise during optimization, the detected result tends to be positioned between potential targets, leading to local optima. Conversely, the RIS task, involving finer-grained pixel-level predictions, can more precisely identify the target but often lacks sufficient spatial awareness. Thus, it becomes essential to introduce a multi-task consistency constraint to guide the model in supplementing information, thereby enhancing recognition in ambiguous situations.
To this end, we propose a coarse-to-fine architecture for multi-task visual grounding, named $\text{C}^3\text{VG}$.
The structure is shown in Fig.~\ref{fig:framework}. Initially, we employ a pixel decoder and a query decoder to independently generate coarse foreground semantics and localization regions in the Rough Semantic Perception (RSP) stage. Subsequently, the Refined Consistency Interaction (RCI) stage refines them and enforces consistency across the multi-task outcomes. Within the RCI stage, we introduce a Mask-guided Interaction Module (MIM) to implicitly integrate the multi-task results from the RSP stage. Furthermore, we apply a bidirectional consistency constraint loss to explicitly enforce consistency across tasks.
As illustrated in Fig.~\ref{fig:motivation2}(a), the RSP stage delivers coarse localization and semantic results. Building on these priors, the RCI stage applies consistency constraints to produce higher-quality predictions.

\textbf{Insufficient multimodal understanding} primarily manifests as an inability to effectively capture the semantic associations between modalities in downstream tasks, particularly when data is limited.
Fig.~\ref{fig:motivation}(b) shows two instances of identification errors caused by inadequate multimodal understanding: (1) The model incorrectly identifies `egg cup' by focusing only on `cup'; (2) The model misinterprets `iMac' due to the absence of prior knowledge.
Recently, SimVG~\cite{simvg} has confirmed the importance of employing a pretrained multi-modality encoder for improving referential understanding. However, this paper aims to further extend this structure from a single detection task to a multi-task learning framework to validate its broader effectiveness.
%
As shown on the left side of Fig.~\ref{fig:motivation2}(a), previous methods typically utilize single-modal pretrained models as feature encoders and rely on limited downstream data to learn vision-language fusion representations. Recently, SimVG~\cite{simvg} has decoupled the downstream multimodal fusion process and incorporated it into upstream pretraining, resulting in significant performance improvements for the REC task. 
Fig.~\ref{fig:motivation2}(b) illustrates the direct integration of the two modalities during the upstream pretraining process, leveraging advances in vision-language pretraining research~\cite{VILT,beit3}. 
This paper extends the conclusions of SimVG~\cite{simvg}, demonstrating that the integration of multimodal pretrained models significantly enhances both convergence speed and accuracy in RIS and multi-task visual grounding tasks.

Our main contributions are summarized as follows:
\begin{enumerate}
    \item We introduce an innovative and efficient coarse-to-fine architecture, $\text{C}^3\text{VG}$, specifically designed for multi-task visual grounding.
    \item We design a mask-guided interaction module and a bidirectional consistency constraint loss to address the challenge of multi-task prediction inconsistency. These components facilitate implicit interaction and provide explicit supervision for multi-task predictions, respectively.
    \item We extend the pretrained multi-modality encoder from a single-task setting to a multi-task joint training framework and validate its impact on addressing the issue of inadequate multimodal understanding.
    \item The proposed $\text{C}^3\text{VG}$ framework significantly outperforms state-of-the-art methods on RefCOCO/+/g datasets for both REC and RIS tasks, while requiring only half or fewer training epochs.
\end{enumerate}

\section{Related Work}

\subsection{Visual Grounding}
\textbf{Referring Expression Comprehension} (REC)~\cite{nmtree, resc, fgvp, scanformer,falip} predicts a bounding box that tightly encompasses the target object in an image based on a referring expression.
\textbf{Referring Image Segmentation} (RIS)~\cite{lavt, coupalign, caris} aims to provide pixel-level localization of a target object in an image based on a referring expression.
\textbf{Multi-task Visual Grounding} seeks to localize and segment referring expressions using a single, integrated model.
MCN~\cite{mcn} introduces a consistency energy maximization loss, which constrains the feature activation maps in both REC and RIS to be similar.
Some Transformer-based methods~\cite{reftr,eevg} seek more comprehensive multimodal modeling approaches to enhance the performance of multi-task visual grounding.
SeqTR~\cite{seqtr} and PolyFormer~\cite{polyformer} employ a sequential transformer model that processes visual and textual data in a unified manner, enhancing performance on multi-task visual grounding by sequentially refining predictions.
Recently, MLLM-based methods~\cite{lisa, gsva} leverage the capabilities of MLLM~\cite{llava,ST3} to enforce rule-based serialization of predictions, effectively integrating the REC and RIS tasks into a unified framework.
Our work follows the paradigm of MCN, which primarily explores and investigates consistency constraints. However, our proposed $\text{C}^3\text{VG}$ further enhances model consistency prediction through implicit interactions and explicit supervision.

\subsection{Vision Language Pre-Training (VLP)}
Existing VLP models can be broadly categorized into three types.
One-stream models~\cite{UNITER,alberf,soho} process both image and text inputs in a single stream. They concatenate image and text embeddings and interact cross-modality information throughout the entire feature extraction process.
Dual-stream models~\cite{CLIP,align,declip} employ separate encoders for each modality. These models do not concatenate modalities at the input level; instead, the interaction between pooled image and text vectors occurs at a shallow layer.
Dual-stream models with fusion encoder~\cite{BLIP, vlmo, flava} combine aspects of both one-stream and dual-stream models. They facilitate intermediate interaction between modalities, potentially striking a balance between complexity and performance.
Visual grounding fundamentally constitutes one of the downstream task of VLP. CRIS~\cite{cris} and Dynamic MDETR~\cite{dynamicmdetr} apply dual-stream vision-language pre-training models to leverage their feature alignment and enhanced modality representation capabilities. SimVG~\cite{simvg} decouples the concept of multimodal mutual understanding from downstream tasks with limited data to the pre-training phase, achieving significant performance improvements in REC tasks. This paper further addresses the issue of insufficient multimodal understanding for multi-task joint training by employing multimodal fusion representations pre-training method~\cite{beit3}.

\section{The Proposed $\text{C}^3\text{VG}$}\label{sec:method}

\begin{figure*}[t]
	\centering
	\includegraphics[width=1.0\linewidth]{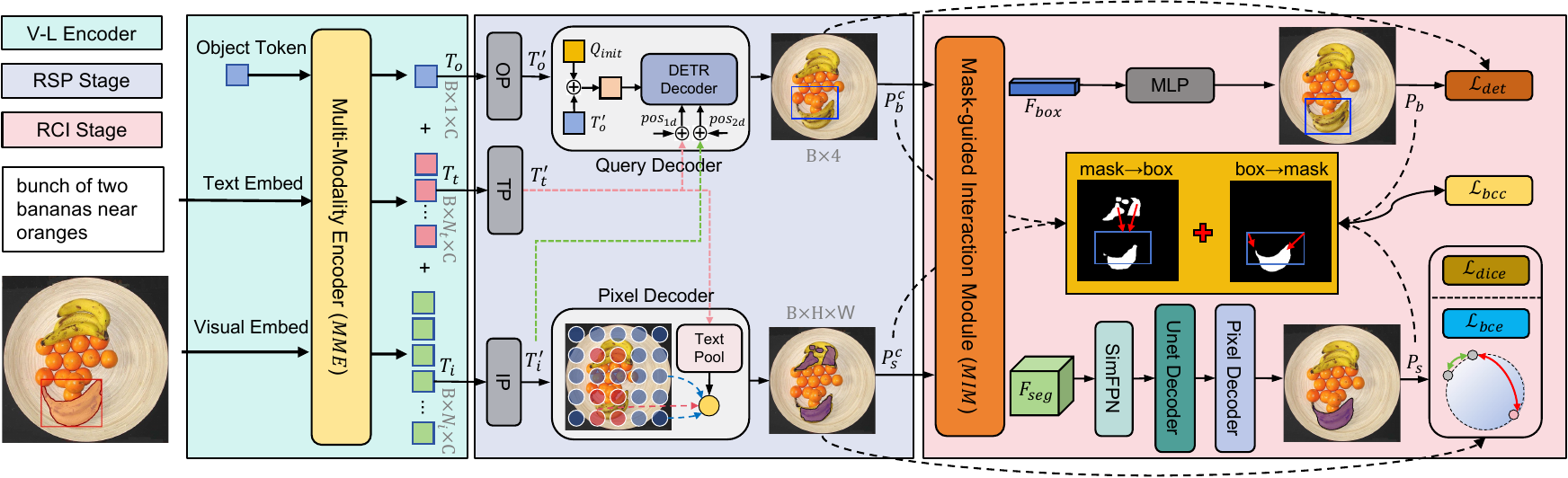} 
	\caption{
		The overall framework of the proposed $\text{C}^3\text{VG}$. First, the image and text features are fused and encoded using a multi-modality encoder. In the RSP stage, the pixel decoder and query decoder generate coarse segmentation and detection results. In the RCI stage, these multi-task priors are further refined through interaction and consistency constraints.
	}
	\label{fig:framework}
\end{figure*}

\subsection{Architecture Overview}\label{subsec:overall}
Fig.~\ref{fig:framework} provides an overview of the $\text{C}^3\text{VG}$ architecture. Initially, the image and text modalities are independently embedded and processed through a multi-modality encoder (MME) for vision-language encoding and fusion, positioning the joint representation of multimodal fusion upstream. A learnable object token is also utilized as the feature representation for the REC task. The framework then advances through the RSP and RCI stages, ultimately yielding high-quality predictions.

\paragraph{Multi-Modality Encoder.}
The input to $\text{C}^3\text{VG}$ consists of an image ${I}\in\mathbb{R}^{3\times{H}\times {W}}$ and a caption text ${T}\in\Omega^{M}$, where $\Omega$ denotes the vocabulary set. The image is initially downsampled to 1/16 of its original size using a visual embedding, resulting in ${P}_{i}=\{p^1,p^2,\dots,p^{N_i}\}$. The text is then tokenized into ${L}_{t} = \{l^1, l^2, \dots, l^{N_t}\}$.
Additionally, we define a learnable object token $T_{o}$ as the target feature for the REC branch. The inputs of MME can be expressed as:
\begin{equation}
	\setlength{\abovedisplayskip}{2pt}
	\setlength{\belowdisplayskip}{2pt}
	\begin{aligned}
		\text{T} = \{T_{o},p^1,p^2\dots,p^{N_i},l^1,l^2,...,l^{N_t}\}. \\
	\end{aligned}
	\vspace{0pt}
	\label{eq_token_mask}
\end{equation}
The MME architecture leverages the pre-trained weights of the BEiT-3~\cite{beit3} model.
The output of the MME comprises three components: $T_o\in\mathbb{R}^{B\times{1}\times {C}}$, ${T_t}\in\mathbb{R}^{B\times{N_t}\times {C}}$, ${T_i}\in\mathbb{R}^{B\times{N_i}\times {C}}$.
\paragraph{Rough Semantic Perception Stage.}
The RSP stage aims to generate a rough localization and semantic outline, serving as priors for the RCI stage.
Initially, the outputs of the MME are projected to a common dimension via three unshared linear layers:
\begin{equation}
	\setlength{\abovedisplayskip}{2pt}
	\setlength{\belowdisplayskip}{2pt}
	\begin{aligned}
		T_o^{'} = \text{OP}(T_o),\ \ \ \ T_t^{'} = \text{TP}(T_t),\ \ \ \  T_i^{'} = \text{IP}(T_i).
	\end{aligned}
	\vspace{0pt}
	\label{eq_qam}
\end{equation}
For the REC branch, the process begins with a query decoder, which enhances the representation of the object token by interacting with text and image tokens. The query decoder is defined as:
\begin{equation}
\setlength{\abovedisplayskip}{2pt}
\setlength{\belowdisplayskip}{2pt}
\begin{aligned}
    T_c = \text{MCA}(\text{MLP}(\text{Concat}(T_o^{'}, \text{MCA}(T_o^{'}+Q_{init}, \\ T_t^{'}+pos_{1d}))), T_i^{'}+pos_{2d}),
\end{aligned}
\vspace{0pt}
\label{eq_qam}
\end{equation}
where MCA($A_1$, $A_2$) denotes the multi-head cross attention mechanism, with $A_1$ serving as the query and $A_2$ as the key and value. Subsequently, an MLP is employed to regress and predict the REC output $P_b^c\in\mathbb{R}^{B\times4}$.
For the RIS branch, we adopt a text-to-pixel correlation strategy similar to CRIS~\cite{cris} to generate the predicted mask $P_s^c\in\mathbb{R}^{B\times H\times W}$. However, instead of using a 3$\times$3 convolution with padding, we compress the text using a 1$\times$1 convolution without additional padding.
\paragraph{Refined Consistency Interaction Stage.}
The Refined Consistency Interaction (RCI) stage is designed to harmonize the outputs from the RSP stage, ensuring multi-task consistency through both implicit interactions and explicit constraints. We first introduce a mask-guided interaction module (MIM) that adaptively and implicitly aligns the consistency between the detection and segmentation predictions. Additionally, an auxiliary bidirectional consistency constraint loss is incorporated to explicitly enforce alignment at the result level.
In the REC branch, an MLP layer is utilized to regress object features at the RCI stage. In the RIS branch, we integrate SimFPN~\cite{vitdet} to capture multi-level structures, followed by a UNet-style~\cite{unet} decoder that performs multi-level fusion and a pixel decoder, consistent with the methodology employed in the RSP stage.

\subsection{Mask-guided Interaction Module}\label{subsec:uim}
\begin{figure}[t]
	\centering
	\includegraphics[width=1.0\linewidth]{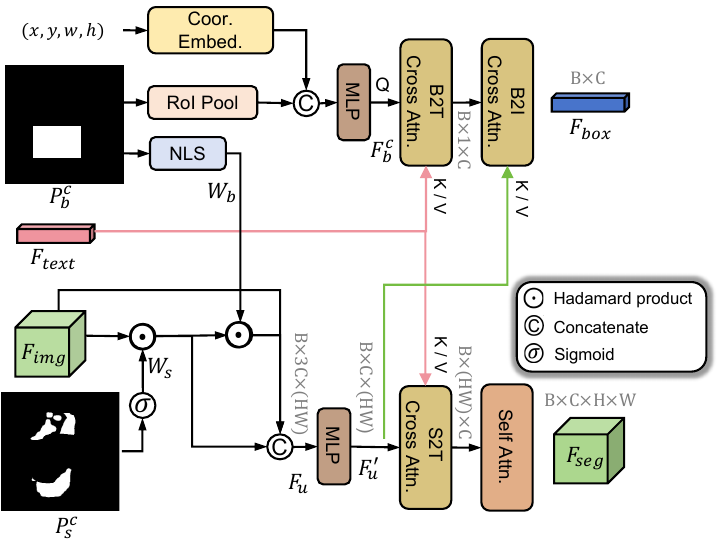} 
	\caption{
  		Architecture of the Mask-guided Interaction Module (MIM). "Coor. Embed" denotes a linear layer that maps coordinate positions into the hidden space.
	}
	\label{fig:uim}
\end{figure}
The RSP stage provides spatial prior information for the RCI stage, while the MIM is designed to implicitly model the relationships between the multi-task results from the RSP stage in a learnable manner.
In the REC branch, based on the detection results from the RSP stage \(P_{b}^{c} \in \mathbb{R}^{B \times 4}\), which are represented as \((x,y,w,h)\), two operations are performed. (1) The results are used as the ROI to pool features from \(F_{img}\). (2) Coordinate representations are obtained through coordinate embedding (CoE). The RSP stage box feature \(F_{b}^c\) is then computed as follows: 
\begin{equation}
	\begin{aligned}
		F_b^c = \text{MLP}(\text{Concat}(\text{RoIP}(P_b^c, F_{img})+\text{CoE}(P_b^c))).
	\end{aligned}
	\vspace{0pt}
	\label{eq_uim_boxpool}
\end{equation}
where RoIP denotes the RoI Pooling operation as in Faster R-CNN~\cite{fasterrcnn}. To enable the bounding box to utilize the structural information from the RIS branch and ensure consistent predictions, we interact \(F_b^c\) with both textual and visual features. The final interacted object feature \(F_{box}\) is expressed as:
\begin{equation}
	\begin{aligned}
		F_{box} = \text{MCA}(\text{MCA}(F_b^c, F_{text}), F_u^{'}),
	\end{aligned}
	\vspace{0pt}
	\label{eq_uim_boxattn}
\end{equation}
where the calculation of \( F_u^{'} \) is detailed in Eq. \ref{eq_uim_fseg}.

\begin{figure}[t]
	\centering
	\includegraphics[width=1.0\linewidth]{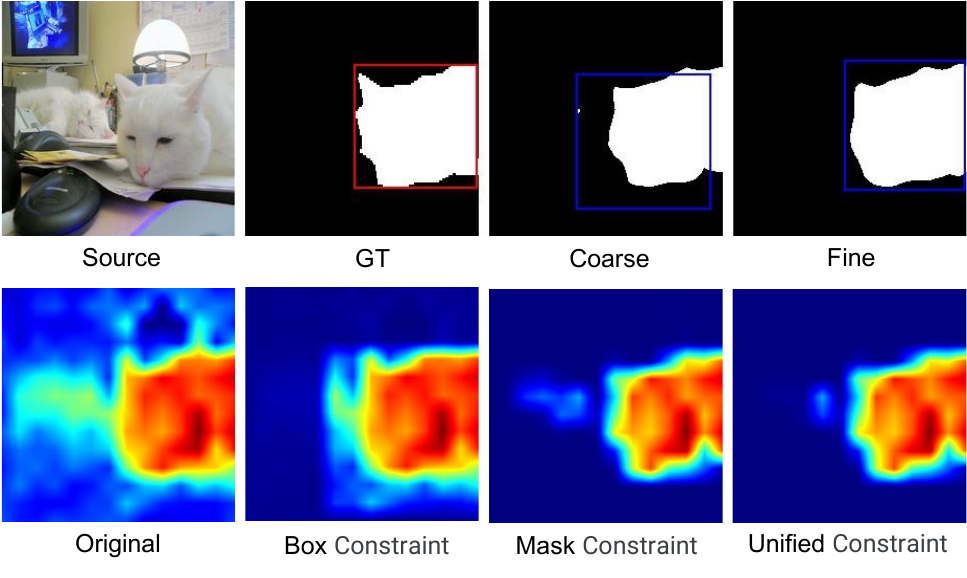} 
	\caption{Visualization of intermediate model processes. First row: original image, GT, RSP stage, and RCI stage results. Second row: original, box-constrained, mask-constrained, and unified-constrained heatmaps.}
	\label{fig:uim_visual}
\end{figure}

In the RIS branch, we apply the concept of background suppression and foreground enhancement by leveraging the results of both the REC and RIS branches on \(F_{img}\). First, \(P_b^c\) is converted to the top-left and bottom-right format by rounding to integers as follows:
\begin{equation}
	\begin{aligned}
		x_1 = (x-0.5w)\times w, \ \ y_1 =(y-0.5h)\times h, \\
		x_2 = (x+0.5w)\times w, \ \ y_2 = (y+0.5h)\times h,
	\end{aligned}
	\vspace{0pt}
	\label{eq_uim_x1y1x2y2}
\end{equation}
\begin{equation}
	\begin{aligned}
		\{x_1^{'}, y_1^{'}, x_2^{'}, y_2^{'}\} =\{\lfloor x_1 \rfloor, \lfloor y_1 \rfloor, \lceil x_2 \rceil, \lceil y_2 \rceil\},
	\end{aligned}
	\vspace{0pt}
	\label{eq_uim_newbox}
\end{equation}
where \( \lfloor * \rfloor \) denotes the floor function, and \( \lceil * \rceil \) denotes the ceiling function. The NLS generates a weight mask \( W_b \) of the same dimensions as \( F_{img} \), calculated as follows:
\begin{equation}
	\begin{aligned}
		W_b =
		\begin{cases}
			w_1, & \text{if } x_i \in [x_1^{'}, x_2^{'}] \land y_j \in [y_1^{'}, y_2^{'}] \\
			1, & \text{otherwise}, 
		\end{cases}
	\end{aligned}
	\vspace{0pt}
	\label{eq_uim_wb}
\end{equation}
where \(\forall x_i \in [0, w]\) and \(\forall y_j \in [0, h]\). \( w_1 \) is set to default values of 0.1, respectively.
We then apply a sigmoid function to the predicted mask from the RSP stage to generate the weighted mask \(W_s = \sigma(P_s^c)\). The weights \(W_b\) and \(W_s\) are applied to \(F_{img}\) to obtain the box and mask-constrained feature \(F_u\):
\begin{equation}
	\begin{aligned}
		F_s =& W_s \odot F_{img}, \\
		F_u = \text{Concat}&(F_s, W_b \odot F_s, F_{img}).
	\end{aligned}
	\vspace{0pt}
	\label{eq_uim_fu}
\end{equation}
Next, an MLP reduces the channel dimension from \(3 \times C\) back to the original \(C\), yielding the fused image representation \(F_u^{'}\), which incorporates the predictions from the RSP stage. This process implicitly provides the RCI stage with prior spatial attention information derived from detection and segmentation predictions.
As illustrated in Fig.~\ref{fig:uim_visual}, the presence of two cats results in divergent attention predictions, leading to suboptimal adjustments of the bounding box prediction during the RSP stage. The MIM mitigates this issue by imposing constraints on the regions of high response within the image space, thereby reducing the model's focus on irrelevant targets and enabling more precise target identification.
Furthermore, the fused image representation is interacted with the text, followed by a multi-head self-attention (MSA) layer to further learn consistent semantic associations. This process is expressed as follows:
\begin{equation}
	\begin{aligned}
		F_u^{'}=& \text{MLP}(F_u), \\
		F_{seg} = \text{MSA}&(\text{MCA}(F_u^{'}, F_{text})).
	\end{aligned}
	\vspace{0pt}
	\label{eq_uim_fseg}
\end{equation}
\subsection{Bidirectional Consistency Constraint Loss}\label{subsec:cc}
To complement the implicit interactions facilitated by the MIM across multi-task outputs, we propose an explicit bidirectional consistency constraint loss, denoted as \({\cal L}_{bcc}\).
First, \({\cal L}_{m2b}\), is designed to enforce the segmentation mask to be contained within the predicted bbox:
\begin{equation}
	\begin{aligned}
		{\cal L}_{m2b} = 1 - \frac{\sum (M_s \odot M_b)}{\sum M_s},
	\end{aligned}
	\vspace{0pt}
	\label{eq_cc_s2b}
\end{equation}
\begin{equation}
	\begin{aligned}
		M_s = 
		\begin{cases}
			1, & \text{if } p_{i,j}^{s} > t \\
			0, & \text{otherwise}
		\end{cases}, 
		M_b =
		\begin{cases}
			1, & \text{if } (x_i,y_j) \in P_b \\
			0, & \text{otherwise}
		\end{cases},
	\end{aligned}
	\vspace{0pt}
	\label{eq_cc_s2b_m}
\end{equation}
where \(p^s_{i,j}\) denotes the pixel values of the predicted segmentation mask after applying the sigmoid function, with \(\forall i \in [0, w]\) and \(\forall j \in [0, h]\). \(t\) is set to 0.5. \(P_b\) represents the bounding box prediction.
Second, the loss term \({\cal L}_{b2m}\) is defined as follows:
\begin{equation}
    \begin{aligned}
        {\cal L}_{b2m} = 1 - \frac{|P_b^s \cap P_b|}{|P_b^s \cup P_b|},
    \end{aligned}
    \label{eq:loss_b2m}
\end{equation}
where \(P_b^s\) represents the minimal bounding box that encloses the segmentation mask \(M_s\), and \(P_b\) denotes the predicted bounding box. 
This loss is quantified using the Intersection over Union (IoU) metric, which measures the degree of overlap between the bounding box derived from the segmentation mask and the predicted bounding box. It ensures that the predicted bounding box encapsulates the segmentation mask as comprehensively as possible.
Finally, the overall consistency constraint loss is defined as ${\cal L}_{bcc} = \lambda_1{\cal L}_{b2m} + \lambda_2{\cal L}_{m2b}$, with the weighting coefficients \(\lambda_1\) and \(\lambda_2\) set to 1 and 3, respectively.

\subsection{Training Objectives}\label{subsec:loss}
The primary optimization loss for the multi-task visual grounding is comprised of two main components: REC and RIS, which are defined as follows:
\begin{equation}
    \begin{aligned}
        {\cal L}_{rec} &= \sigma_{l1}{\cal L}_{l1} + \sigma_{giou}{\cal L}_{giou}, \\
        {\cal L}_{ris} &= \sigma_{dice} {\cal L}_{dice} + \sigma_{bce} {\cal L}_{bce},
    \end{aligned}
    \vspace{0pt}
    \label{eq_Lrec}
\end{equation}
where the weighting factors \(\sigma_{l1}\) and \(\sigma_{giou}\) are set to 0.5 and 0.2, respectively, while \(\sigma_{dice}\) and \(\sigma_{bce}\) are both set to 1.0 by default.
Both ${\cal L}_{rec}$ and ${\cal L}_{ris}$ include two-stage components and are augmented by the bidirectional consistency constraint loss, ${\cal L}_{bcc}$. The total loss is formulated as:
\begin{equation}
	\setlength{\abovedisplayskip}{2pt}
	\setlength{\belowdisplayskip}{2pt}
	\begin{aligned}
		{\cal L}_{total} = \lambda_{c}({ \lambda_{rec}\cal L}_{rec}^{c} + {\cal L}_{ris}^{c}) + \\ ( \lambda_{rec}{\cal L}_{rec}^{f} + {\cal L}_{ris}^{f}) + \lambda_{bcc}{\cal L}_{bcc}
	\end{aligned}
	\vspace{0pt}
	\label{eq_cc_b2s}
\end{equation}
where \(\lambda_{rec}\), \(\lambda_{bcc}\), and \(\lambda_{c}\) are set to 0.5, 0.1, and 0.3, respectively. Here, \({\cal L}_{rec}^{c}\) denotes the REC loss in the RSP stage, while \({\cal L}_{ris}^{f}\) corresponds to the RIS loss in the RCI stage.

\section{Experiments}\label{sec:experiment}
%

\subsection{Experimental Setup}\label{subsec:implementation}
We evaluate the proposed model in RefCOCO~\cite{refcoco/+}, RefCOCO+ and RefCOCOg~\cite{refcocogumd} datasets. 
The maximum sentence length is set to 20. The images are resized to $320\times 320$.
Based on previous works \cite{seqtr}, mIoU and Prec@0.5(Acc(REC) in ablation study) are adopted to evaluate the performance of methods. 
We train our models for 30 epochs with a batch size of 16. Adam~\cite{adam} is adopted as our optimizer.
All experiments are conducted on a system with dual NVIDIA 4090 GPUs.
Further details will be provided in the supplementary materials.

\begin{table*}[t]
	\centering
	\renewcommand{\arraystretch}{0.95}
	\setlength{\tabcolsep}{1.8mm}
	\scriptsize
	\begin{tabular}{l|c|c|c|ccc|ccc|cc|c}
		\noalign{\hrule height 1.5pt}
		\multicolumn{1}{c|}{\multirow{2}{*}{Method}} & \multicolumn{1}{c|}{\multirow{2}{*}{Publication}} & \multicolumn{1}{c|}{\multirow{2}{*}{Backbone}} & \multicolumn{1}{c|}{\multirow{2}{*}{Data Size}} &
		\multicolumn{3}{c|}{RefCOCO} & \multicolumn{3}{c|}{RefCOCO+} & \multicolumn{2}{c|}{RefCOCOg} & \multicolumn{1}{c}{\multirow{1}{*}{Time}}\\
		\cline{5-12}
		& &  &  & val & test A & test B & val & test A & test B & val(U) & test(U) &  (ms)\\
		\hline
		\rowcolor{ggray!20}
		\multicolumn{13}{c}{\it{Single-task}}\\
		\hline
		MDETR~\cite{mdetr} & ICCV2021 & EfficientNet-B3 & 200K& 86.75 & 89.58 & 81.41 & 79.52 & 84.09 & 70.62 & 81.64 & 80.89 & 108 \\
		TransVG++~\cite{transvg++} & T-PAMI2023 & ViT-B & - &86.28 &88.37& 80.97 &75.39 &80.45 &66.28& 76.18 &76.30&- \\
		Dyn.MDETR~\cite{dynamicmdetr} & T-PAMI2023 & ViT-B & - & 85.97 & 88.82& 80.12 &74.83 &81.70 &63.44 &72.21 & 74.14 & -\\
		GroundingDINO~\cite{groundingdino} & ECCV2024 & Swin-T &200K&  89.19 & {91.86} & 85.99 & 81.09 &87.40 &74.71 & 84.15& 84.94& 120 \\
		SimVG~\cite{simvg} & NeurIPS2024 & BEiT3-ViT-B &174K&  \underline{90.59} & {92.80} & 87.04 & {83.54} & {88.05} & {77.50} & {85.38} & \underline{86.28} & \textbf{44} \\
		\hline
		\rowcolor{ggray!20}
		\multicolumn{13}{c}{\it{Multi-task}}\\
		\hline
		MCN~\cite{mcn} & CVPR2020 &DarkNet53 & -& 80.08 & 82.29 & 74.98 & 67.16 & 72.86 & 57.31 & 66.46 & 66.01 & 56 \\
		SeqTR~\cite{seqtr} &ECCV2022  & DarkNet53 & 174K& 81.23 & 85.00 & 76.08 & 68.82 & 75.37 & 58.78 & 71.35 & 71.58 &  \underline{50} \\
		PolyFormer~\cite{polyformer} & CVPR2023 & Swin-B & 174K& {89.73} & 91.73 & {86.03} & {83.73} & {88.60} & {76.38} & {84.46} & {84.96} & 152 \\
		PVD~\cite{pvd} & AAAI2024 & Swin-B & - & 84.52& 87.64 & 79.63 & 73.89 & 78.41 & 64.25 & 73.81 & 74.13& - \\
		EEVG~\cite{eevg} & ECCV2024 & ViT-B & 174K & 90.47 & 92.73 & \underline{87.72} & 81.79 & 87.80 & 74.94 & 85.19 & 84.72 & 117\\ 
		\hline
		\rowcolor{ggray!20}
		\multicolumn{13}{c}{\it{Generalist Models}}\\
		\hline
		Ferret~\cite{ferret} & ICLR2024 & Vicuna-7B & $>8\text{M}$ &87.49& 91.35& 82.45& 80.78& 87.38& 73.14& 83.93& 84.76& - \\
		LION-12B~\cite{lion} & CVPR2024 & FlanT5-11B & 3.6M &  89.80 & \underline{93.02} & 85.57 & \underline{83.95} &\underline{89.22} &\underline{78.06} & \underline{85.52} & 85.74 & - \\
		\hline
		\rowcolor{aliceblue!60}  $\text{C}^3\text{VG}$ & AAAI2025 & BEiT3-ViT-B & 28K & \textbf{92.51} & \textbf{94.60} & \textbf{88.71} & \textbf{87.44} & \textbf{90.69} & \textbf{81.42} & \textbf{87.68} & \textbf{88.31} & {51} \\
		\noalign{\hrule height 1.5pt}
	\end{tabular}
	\caption{\textbf{Main results} on REC datasets. \textbf{Bold} denotes the best performance. \underline{Underline} denotes the second best performance.
	}
	\label{tab:main_rec}
\end{table*}
\begin{table*}[t]
	\centering
	\renewcommand{\arraystretch}{0.95}
	\setlength{\tabcolsep}{2.0mm}
	\scriptsize
	\begin{tabular}{l|c|c|c|c|ccc|ccc|cc}
		\noalign{\hrule height 1.5pt}
		\multicolumn{1}{c|}{\multirow{2}{*}{Method}} &\multicolumn{1}{c|}{\multirow{2}{*}{Publication}} & \multicolumn{1}{c|}{\multirow{2}{*}{Backbone}} & \multicolumn{1}{c|}{\multirow{2}{*}{Data}} & \multicolumn{1}{c|}{\multirow{2}{*}{FT}} &
		\multicolumn{3}{c|}{RefCOCO} & \multicolumn{3}{c|}{RefCOCO+} & \multicolumn{2}{c}{RefCOCOg}\\
		\cline{6-13}
		& & &  & & val & test A & test B & val & test A & test B & val(U) & test(U)  \\ 
		\hline
		\rowcolor{ggray!20}
		\multicolumn{13}{c}{\it{Single-task}}\\
		\hline
		CRIS~{\cite{cris}} &CVPR2022 & ResNet101 & RefC & \xmark  & 70.47 & 73.18 & 66.10 & 62.27 & 68.06 & 53.68 & 59.87 & 60.36\\
		LAVT~\cite{lavt} & CVPR2022& Swin-B & RefC & \xmark  &{74.46} & {76.89} & {70.94} & {65.81} & {70.97} & {59.23} & {63.34} & {63.62}\\
		ReLA~\cite{gres} & CVPR2023 & Swin-B & RefC & \xmark  & 73.82 & 76.48 & 70.18 & 66.04 & 71.02 & 57.65 & 65.00 & 65.97 \\
		Prompt-RIS~{\cite{prompt-ris}} & CVPR2024 & CLIP-ViT-B & Com-RefC & - & 78.10 & 81.21 & 74.64 & 71.13& 76.60 & 64.25 & 70.47 & 71.29 \\
		OneRef~\cite{oneref} & NeurIPS2024 & BEiT3-ViT-B & Com-RefC & \cmark &  \underline{79.83} & \underline{81.86} & 76.99 & \underline{74.68} & \underline{77.90} & \underline{69.58} & \underline{74.06} & \underline{74.92} \\
		\hline
		\rowcolor{ggray!20}
		\multicolumn{13}{c}{\it{Multi-task}}\\
		\hline
		MCN~{\cite{mcn}} & CVPR2020 & DarkNet53 & RefC & \xmark  & 62.44 & 64.20 & 59.71 & 50.62 & 54.99 & 44.69 & 49.22 & 49.40 \\
		SeqTR~{\cite{seqtr}} & ECCV2022 & DarkNet53 & Com-RefC & \cmark & 71.70 & 73.31 & 69.82 & 63.04 & 66.73 & 58.97 & 64.69 & 65.74 \\
		PolyFormer~{\cite{polyformer}} &CVPR2023& Swin-B & Com-RefC & \cmark & 75.96 & 77.09 & 73.22 & 70.65 & 74.51 & 64.64 & 69.36 &69.88 \\
		PVD~\cite{pvd} & AAAI2024 & Swin-B & Com-RefC & \cmark & 74.82& 77.11 &  69.52 & 63.38 & 68.60 & 56.92 & 63.13 & 63.62\\
		EEVG~\cite{eevg} & ECCV2024 & ViT-B & Com-RefC & - & 79.49 & 80.87 & \underline{77.39} & {71.86} & {76.67} & {66.31} & {73.56} & {73.47} \\       
		\hline
		\rowcolor{ggray!20}
		\multicolumn{13}{c}{\it{Generalist Models}}\\
		\hline
		LISA~\cite{lisa} & CVPR2024 & Vicuna-7B & - & \cmark & 74.90 & 79.10 & 72.30 & 65.10 & 70.80 & 58.10 & 67.90 & 70.60 \\
		GSVA~\cite{gsva} & CVPR2024 & Vicuna-7B & - & \cmark & 77.20 & 78.90 & 73.50 & 65.90& 69.60 &59.80 &72.70 & 73.30 \\
		\hline
		\rowcolor{aliceblue!60}  $\text{C}^3\text{VG}$ & AAAI2025 & BEiT3-ViT-B & Com-RefC & \xmark & \textbf{81.37} & \textbf{82.93} & \textbf{79.12} & \textbf{77.05} & \textbf{79.61} & \textbf{72.40} & \textbf{76.34} & \textbf{77.10} \\
		\rowcolor{aliceblue!60}  $\text{C}^3\text{VG}$-oIoU & AAAI2025 & BEiT3-ViT-B & Com-RefC & \xmark  & 80.89 &83.18 & 77.86 & 74.68 & 77.96 & 68.95& 74.43 & 76.39 \\
		\noalign{\hrule height 1.5pt}
	\end{tabular}
	\caption{\textbf{Main Results} on RIS Datasets. \textbf{Bold} indicates the best performance, and \underline{underline} indicates the second-best performance. RefC represents training on a single dataset, while Com-RefC refers to the union of the RefCOCO, RefCOCO+, and RefCOCOg training sets. FT denotes whether fine-tuning is performed on the specific dataset.}
	\label{tab:main_res}
\end{table*}

\subsection{Main Results}\label{subsec:mainresults}
\noindent\textbf{Referring Expression Comprehension.} The single-task part presented in Tab.~\ref{tab:main_rec} showcase a comparison between our method and prior advanced REC approaches.
In comparison to Dynamic MDETR, which utilizes ViT-B as its backbone, $\text{C}^3\text{VG}$ achieves a remarkable improvement of +5.78\%-17.98\% in Acc(REC). Furthermore, when compared to GroundingDINO~\cite{groundingdino}, which is trained on large-scale data, $\text{C}^3\text{VG}$ delivers a gain of +2.72\%-6.71\% in Acc(REC) while also reducing inference latency by 58\%.

\noindent\textbf{Referring Image Segmentation.} The single-task part presented in Tab.~\ref{tab:main_res} compare our $\text{C}^3\text{VG}$ with previous advanced RIS methods.
Our $\text{C}^3\text{VG}$ demonstrates an absolute improvement of 9.75\%-18.72\% over the Transformer-based CRIS~\cite{cris} model. Additionally, it achieves +1.72\%-8.15\% in mIoU compared to the latest SOTA model Prompt-RIS~\cite{prompt-ris}, under the same ViT-B backbone conditions.

\noindent\textbf{Multi-Task Visual Grounding.} The multi-task results presented in Tab.~\ref{tab:main_rec} and Tab.~\ref{tab:main_res} provide a comparative analysis between the proposed $\text{C}^3\text{VG}$ and existing multi-task visual grounding approaches. 
Compared to PolyFormer~\cite{polyformer}, our $\text{C}^3\text{VG}$ demonstrates marked improvements, surpassing it by margins of +2.09\%-5.04\% in Acc(REC) and +5.10\%-7.76\% in mIoU. Furthermore, our method exhibits inference efficiency comparable to that of SeqTR, nearing real-time performance.

\noindent\textbf{Generalist Models.} 
Multimodal Large Language Models~\cite{jin2024emllm} have also expanded into the visual grounding domain, with their results listed under the generalist models part in Tab.~\ref{tab:main_rec} and Tab.~\ref{tab:main_res}. These models are distinguished by enormous parameters and extensive pretraining on vast datasets, providing strong generalization capabilities. However, our method demonstrates strong competitiveness compared to these generalist models.

\subsection{Ablation Studies}
\label{subsec:quan_ana}
\noindent\textbf{Basic Improvement Setting.} 
We implement several techniques to enhance the performance of our baseline model, with the experimental outcomes presented in Tab.~\ref{tab:ab_overall}. The baseline architecture leverages the ViT-B and BERT models as the visual and textual encoders, respectively, with VGTR head.
First, we observe a substantial performance boost by incorporating multimodal fusion representation pretraining (BEiT-3), which yields an increase of +5.11\% in Acc(REC) and +5.28\% in oIoU.
This improvement can be attributed to the fact that prior methods often rely on limited downstream data to learn multimodal representations, resulting in inadequate multimodal comprehension. Given the complex and rich semantics inherent in text, pretraining multimodal representations is essential for achieving sophisticated multimodal understanding. 
%
Furthermore, the joint training of REC and RIS has shown a mutually beneficial effect, leading to an improvement of +1.68\% in Acc(REC) and +1.47\% in oIoU. Finally, the integration of SimFPN, which facilitates comprehensive interaction across multi-level features, further enhances oIoU by an additional +1.05\%.
\begin{table}
	\centering
	\renewcommand{\arraystretch}{1.0}
	\setlength{\tabcolsep}{2.4mm}
	\scriptsize
	\begin{tabular}{l|ccc}
		\noalign{\hrule height 1.5pt}
		Method & Acc (REC) & Acc (RIS) & oIoU (RIS) \\
		\hline
		Baseline & 75.33 & 74.21 & 62.21 \\
            \hline
            \rowcolor{aliceblue} + MM Pretrain& 80.44~\increase{5.11} & 80.08~\increase{5.87} & 67.49~\increase{5.28} \\
		\rowcolor{aliceblue} + Multi-Task& 82.12~\increase{1.68} & 81.78~\increase{1.70} & 68.96~\increase{1.47} \\
		\rowcolor{aliceblue} + SimFPN & 82.25~\increase{0.13} & 82.42~\increase{1.36} &70.01~\increase{1.05} \\
            \hline
		\rowcolor{aliceblue} +  Query Decoder & 84.51~\increase{2.26} & 82.12~\deincrease{0.30} & 69.81~\deincrease{0.20} \\
            \rowcolor{aliceblue} +  Pixel Decoder & 84.35~\deincrease{0.16} & 83.23~\increase{1.11} & 70.81~\increase{1.00} \\
        \noalign{\hrule height 1.5pt}
	\end{tabular}
	\caption{Ablation study on basic improvement settings.}
	\label{tab:ab_overall}
\end{table}

\noindent\textbf{Query / Pixel Decoder.} 
The query decoder is designed to integrate guidance from both textual and visual modalities into the tokens utilized by the detection branch, thereby improving localization accuracy. As demonstrated in Tab.~\ref{tab:ab_overall}, the incorporation of the query decoder leads to a +2.26\% increase in Acc(REC). The pixel decoder, on the other hand, estimates the confidence of each pixel belonging to the foreground through text-pixel contrastive learning. This addition strengthens the supervision within the segmentation branch, resulting in a +1.00\% enhancement in oIoU.

\noindent\textbf{Consistency Constraint Loss.}
This paper introduces two directions of consistency constraint losses for optimization: mask$\rightarrow$box (${\cal L}_{m2b}$) and box$\rightarrow$mask (${\cal L}_{b2m}$). The purpose of ${\cal L}_{m2b}$ is to align the RIS-predicted mask distribution with the REC-predicted bounding box. In contrast, ${\cal L}_{b2m}$ is designed to ensure that the REC-predicted bounding box encompasses the RIS-predicted mask while concurrently suppressing extraneous predictions in non-relevant regions.
As demonstrated in Tab.~\ref{tab:ab_cc}, both the ${\cal L}_{m2b}$ and ${\cal L}_{b2m}$ constraints positively influence performance in both REC and RIS tasks.
Moreover, integrating them to establish bidirectional consistency constraints results in further performance enhancements, yielding +1.20\% in Acc(REC) and +1.95\% in oIoU.

\begin{figure}
	\centering
	\begin{minipage}[b]{0.22\textwidth}
		\centering
		\includegraphics[width=\textwidth]{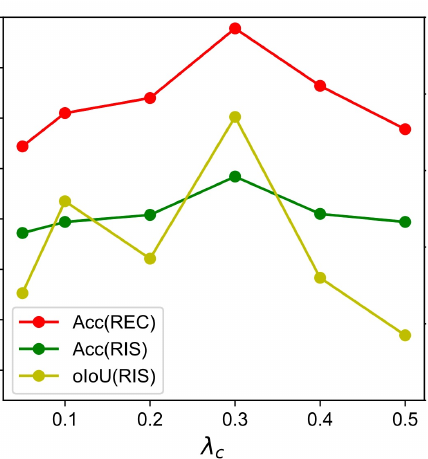}
		\caption{Ablation study on weight of RSP stage $\lambda_c$.}
		\label{fig:lambda_c}
	\end{minipage}
    \hspace{5pt}
	\begin{minipage}[b]{0.215\textwidth}
		\centering
		\includegraphics[width=\textwidth]{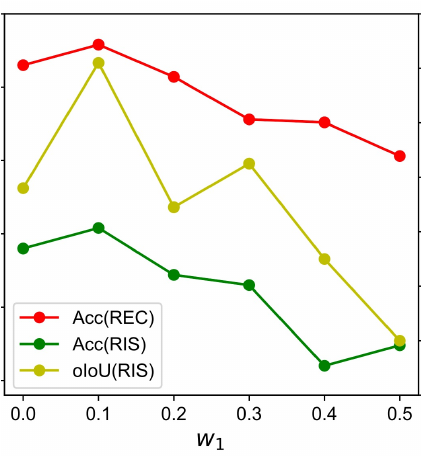}
		\caption{Ablation study on $w_1$ in MIM.}
		\label{fig:w_1}
	\end{minipage}
\end{figure}

\begin{table}
	\centering
	\renewcommand{\arraystretch}{1.0}
	\setlength{\tabcolsep}{2.6mm}
	\scriptsize
	\begin{tabular}{cc|ccc}
		\noalign{\hrule height 1.5pt}
		${\cal L}_{b2m}$ & ${\cal L}_{m2b}$ & Acc (REC) & Acc (RIS) & oIoU (RIS) \\
		\hline
		& & 84.35 & 83.23 & 70.81 \\
            \hline
		\cmark & & 85.12~\increase{0.77} & 83.96~\increase{0.73} & 72.13~\increase{1.32} \\
		& \cmark & 85.01~\increase{0.76} & 84.38~\increase{1.15} & 72.24~\increase{1.43} \\
		\rowcolor{aliceblue}
		\cmark & \cmark & \textbf{85.55~\increase{1.20}} & \textbf{84.59~\increase{1.36}} & \textbf{72.74~\increase{1.93}} \\
		\noalign{\hrule height 1.5pt}
	\end{tabular}
	\caption{Ablation study on consistency constraint loss.}
	\label{tab:ab_cc}
\end{table}
\begin{table}
	\centering
	\renewcommand{\arraystretch}{1.0}
	\setlength{\tabcolsep}{3.0mm}
	\scriptsize
	\begin{tabular}{l|ccc}
		\noalign{\hrule height 1.5pt}
		Method & Acc (REC) & Acc (RIS) & oIoU (RIS) \\
		\hline
		RSP Stage & 84.30 & 83.00 & 70.48 \\
		RCI Stage & 84.54~\increase{0.34} & 84.15~\increase{1.15} & 71.95~\increase{1.47} \\
		\hline
		\rowcolor{ggray!20}
		\multicolumn{4}{l}{\it{REC branch}}\\
		\hline
		\rowcolor{aliceblue}{+Text Attn.} & 85.14~\increase{0.60} & 84.09~\deincrease{0.06} &71.83~\deincrease{0.12} \\
		\rowcolor{aliceblue}{+Coor. Embed} & 85.41~\increase{0.27} & 84.08~\deincrease{0.01} & 71.98~\increase{0.15} \\
		\hline
		\rowcolor{ggray!20}
		\multicolumn{4}{l}{\it{Interaction type}}\\
		\hline
		\multicolumn{1}{r|}{Box } &85.93~\increase{0.52} & 84.22~\increase{0.14} & 71.89~\deincrease{0.09} \\
		\multicolumn{1}{r|}{Mask}& 85.31~\deincrease{0.10} & 84.63~\increase{0.55} &72.71~\increase{0.73} \\
		\rowcolor{aliceblue}\multicolumn{1}{r|}{Unified}& \textbf{86.05~\increase{0.64}} & \textbf{84.80~\increase{0.72}} & \textbf{72.98~\increase{1.00}} \\
		\noalign{\hrule height 1.5pt}
	\end{tabular}
	\caption{Ablation study on unified interaction module. 
}
	\label{tab:ab_uim}
\end{table}

\noindent{\textbf{Mask-guided Interaction Module.}}

As illustrated in Tab.~\ref{tab:ab_uim}, MIM introduces a coarse-to-fine learning paradigm, where the RCI stage demonstrates significant improvements over the RSP stage, particularly in segmentation-related metrics. Moreover, the integration of text interaction further enhances Acc(REC) by +0.60\%, with minimal impact on RIS metrics. The term `Coor. Embed.' pertains to encoding the RSP stage's prediction results $(x, y, w, h)$, which results in a 0.27\% increase in Acc(REC).
In the RIS branch, we conducted ablation studies to assess the introduction of various prior information from the coarse stage, as detailed in the `Interaction type' section of Tab.~\ref{tab:ab_uim}. These studies reveal that incorporating box interaction further strengthens the REC branch. This enhancement is attributed to the interaction between two stages, wherein the RCI stage imposes more stringent requirements on the prediction box generated by the RSP stage. Additionally, the effect of the background weight $w_1$ in NLS is depicted in Fig.~\ref{fig:w_1}, with $w_1=0.1$ employed as the default value in this study.
Similarly, utilizing the mask prior from the RSP stage further improves segmentation performance in the RCI stage.
Finally, unified interaction improves performance by concurrently integrating positional and semantic priors from the RSP stage. By leveraging the complementary information from both sources, it constructs a consistent multi-task representation. As evidenced by the visualization in Fig.~\ref{fig:uim_visual}, this implicit constraint functions as a foreground feature extraction mechanism. Unlike the post-processing employed in MCN~\cite{mcn}, MIM utilizes an implicit, learnable modeling approach to interact with multi-task results, thereby achieving consistent representations.
Fig.~\ref{fig:lambda_c} illustrates the impact of the weight proportion of the coarse stage on the loss calculation. Ultimately, $\lambda_c = 0.3$ is adopted as the default value.

\section{Conclusion}
In this paper, we present $\text{C}^3\text{VG}$, a coarse-to-fine architecture designed for multi-task visual grounding, aimed at addressing issues of prediction inconsistency and inadequate multimodal comprehension.
Initially, during the Rough Semantic Perception (RSP) stage, we extract coarse spatial locations and semantic boundaries using query and pixel decoders. Subsequently, we introduce a mask-guided interaction module to implicitly refine predictions from the RSP stage, while a bidirectional consistency constraint loss explicitly enforces coherence during the Refined Consistency Interaction (RCI) stage.
Furthermore, to address the challenge of insufficient multimodal understanding, we validate the effectiveness of extending the multimodal encoder from a single-task setting to a multi-task joint training framework.
Empirical evaluations substantiate the efficacy and soundness of $\text{C}^3\text{VG}$, which outperforms the existing advanced REC and RIS methods by a remarkable margin.

\section*{Acknowledgments}
This work is supported by the National Natural Science Foundation of China under Nos. 62276061 and 62436002. This work is also supported by Research Fund for Advanced Ocean Institute of Southeast University (Major Program MP202404).

\bibliography{C3VG}

\clearpage

\appendix
\section*{Appendix}

\section{Additional Dataset Details}
\paragraph{\textbf{RefCOCO/RefCOCO+}:} 
RefCOCO comprises 142,209 annotated expressions corresponding to 50,000 objects across 19,994 images, while RefCOCO+ includes 141,564 expressions for 49,856 objects in 19,992 images. Both datasets are divided into training, validation, test A, and test B sets. Test A contains images with multiple people, whereas test B features images with multiple instances of various other objects. Unlike RefCOCO, RefCOCO+ prohibits the use of location-based words in the referring expressions, thus increasing the task's difficulty.

\paragraph{\textbf{RefCOCOg}:}
The RefCOCOg dataset was curated using Amazon Mechanical Turk, where workers were instructed to generate natural language referring expressions for specific objects. It comprises 85,474 referring expressions for 54,822 objects across 26,711 images. Compared to RefCOCO and RefCOCO+, RefCOCOg features longer and more complex expressions, averaging 8.4 words, versus 3.5 words in the other datasets, thereby increasing the challenge. We adopt the UMD partition for RefCOCOg, as it provides distinct validation and testing sets without overlap between training and validation images.

\section{Additional Implementation Details}
In the $\text{C}^3\text{VG}$ model, the output of the original ViT-B is uniformly reduced from 768 to 256 dimensions for subsequent head operations. Specifically, the OP, TP, and IP map features from 768 dimensions to 256. For evaluation metrics, Acc (REC) refers to the accuracy when the box IoU exceeds 0.5, while Acc (RIS) pertains to the accuracy when the mask IoU exceeds 0.5. All experiments are conducted without utilizing the Exponential Moving Average (EMA) technique.
The initial learning rate for the V-L encoder is set at 5e-5, with other parameters at 5e-4. The learning rate undergoes a decay by a factor of 0.1 at the 25th epoch. To ensure a comprehensive presentation of the results, both mIoU and oIoU metrics are included in the SOTA table.
All ablation studies are conducted at a resolution of 224$\times$224, with training over 20 epochs, and the learning rate decays by a factor of 0.1 at the 15th epoch. Metrics are based on the testB split of the RefCOCO dataset. The results in Tab.~\ref{tab:main_rec} and \ref{tab:main_res} are obtained using the combined training data from the unc set of RefCOCO and RefCOCO+, along with the umd set of RefCOCOg.

\section{Additional Method}
\subsection{Decoder Architecture}
The extension of the ViT structure to generate multi-scale feature maps is initially proposed in ViTDet~\cite{vitdet}, termed SimFPN. In our work, we adopt this design to extend the single-scale feature map of the original ViT, resulting in four scales: ${\mathcal{M}_0, \mathcal{M}_1, \mathcal{M}_2, \mathcal{M}_3}$, corresponding to ($\frac{1}{4}$, $\frac{1}{8}$, $\frac{1}{16}$, $\frac{1}{32}$) of the original image, respectively. We then employ a UNet-type decoder to further process these multi-scale features. The UNet-type decoder is described as follows:
\begin{equation}
  \left\{
  \begin{aligned}
  \mathcal{M}_{0}' &= \mathcal{M}_0 , \ \ \ \ \quad i \in \{0, 1, 2\}
   \\
   \mathcal{M}_{i}^{o} &= \text{ConvModule}^{2}(\mathcal{M}_{i}') , \\
   \mathcal{M}_{i+1}' &= \text{Concat}(\mathrm{Up}(\mathcal{M}_{i}^o), \mathcal{M}_{i+1}), 
  \end{aligned}
  \right.
\end{equation}
Here, the ConvModule consists of Convolution, BatchNorm, and ReLU operations. ConvModule$^2$ indicates the repeated application of the ConvModule operations. The final output of the decoder is $\mathcal{M}_3^o$.

\section{Additional Experiments}
\subsection{Ablation Study on Convergence Speed} 
\begin{figure}[t]
	\centering
	\includegraphics[width=1.0\linewidth]{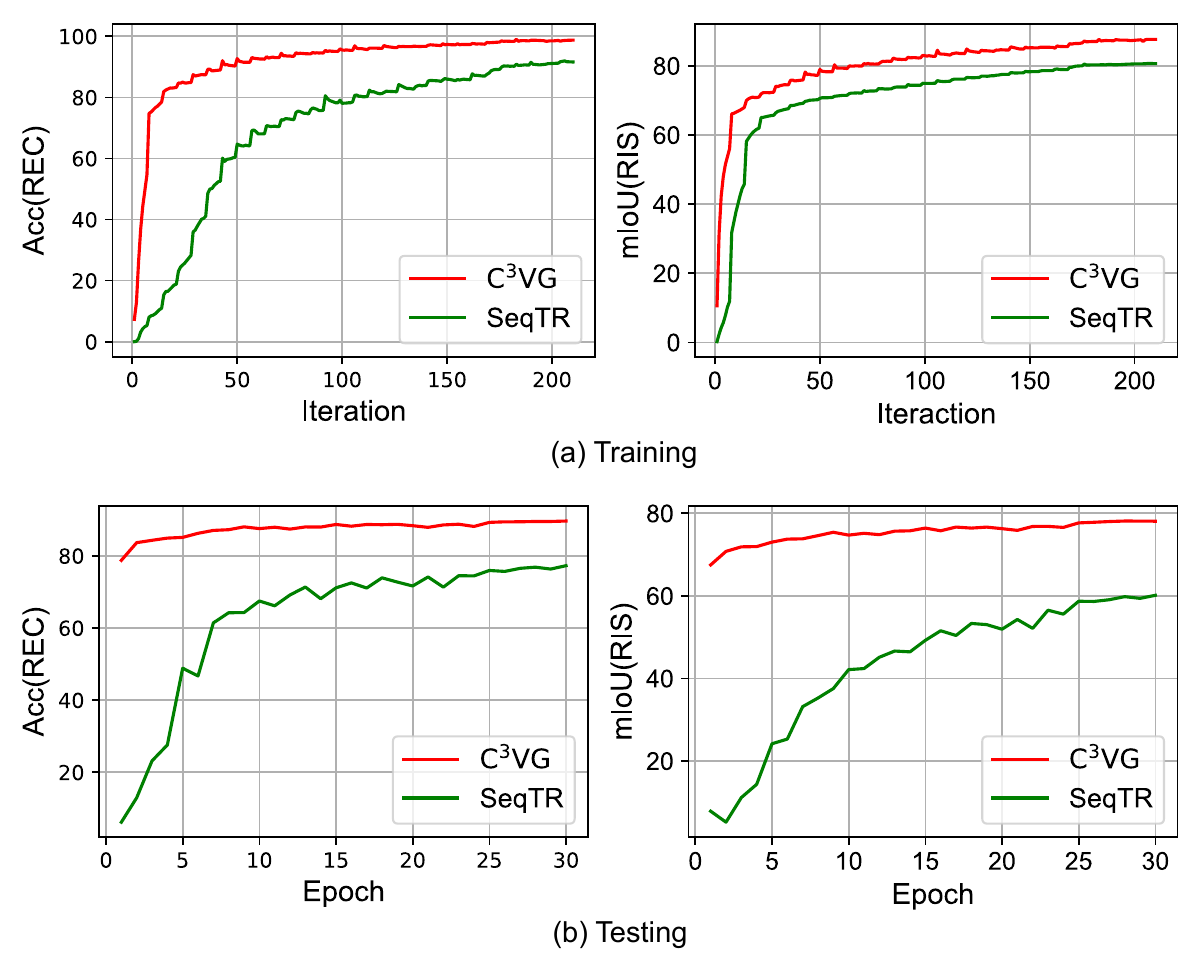} 
	\caption{Comparison of the convergence of $\text{C}^3\text{VG}$ and SeqTR during training and testing. (a) shows Acc (REC) and mIoU (RIS) metrics calculated in the training set, with iterations extracted at regular intervals. (b) shows validation performance after each epoch.}
	\vspace{-10pt}
	\label{fig:converage_speed}
\end{figure}

The $\text{C}^3\text{VG}$ model demonstrates a significantly accelerated convergence speed, attributed to the integration of the Multi-Modality Encoder (MME). We compare the convergence speed of $\text{C}^3\text{VG}$ with SeqTR, as shown in Fig.~\ref{fig:converage_speed}. Panel (a) presents the results from the training set, with iterations sampled at regular intervals, while panel (b) illustrates the validation set performance after each epoch. The proposed method requires substantially fewer epochs (approximately 30) to surpass the performance of existing models, which typically need 60 or more epochs.

\subsection{Parameters of Different Modules}
The $\text{C}^3\text{VG}$ model consists of several modules, each with distinct parameter counts. As shown in Tab.~\ref{tab:exp_param}, the MME module contains approximately 170.6M parameters. The RCP stage (DETR Decoder and Pixel Decoder) includes around 2M parameters, while the MIM module has approximately 5M parameters. SimFPN incorporates 3.6M parameters, and the Unet Decoder comprises 5.2M parameters. Lastly, the detection branch in the RCI stage uses only a small MLP with about 0.1M parameters.

\begin{table}
	\centering
	\renewcommand{\arraystretch}{1.0}
	\setlength{\tabcolsep}{3.0mm}
	
	\begin{tabular}{l|c}
		\noalign{\hrule height 1.5pt}
		Module & Params (M)\\
		\hline
		MME & 170.6 \\
		+RCP stage & 172.5 \\
		+MIM & 177.4 \\
		+SimFPN & 181.0 \\
		+Unet Decoder & 186.2 \\
		\hline
		Total & 186.3 \\
		\noalign{\hrule height 1.5pt}
	\end{tabular}
	\vspace{-5pt}
	\caption{Parameters of different modules. 
}
	\vspace{-5pt}
	\label{tab:exp_param}
\end{table}

\subsection{Comparison of Training Epochs}
We compare the number of epochs required for training $\text{C}^3\text{VG}$ with existing methods in Tab.~\ref{tab:train_epoch}. The proposed method requires only 30 epochs for pre-training and 0 epochs for fine-tuning, significantly fewer than the 60-180 epochs required by existing methods. This accelerated convergence is attributed to the integration of the MME module, which facilitates rapid convergence by leveraging pre-trained multimodal representations.

\begin{table}
	\centering
	\renewcommand{\arraystretch}{1.0}
	\setlength{\tabcolsep}{3.0mm}
	\begin{tabular}{l|c}
		\noalign{\hrule height 1.5pt}
		Method & Epochs\\
		\hline
		\multicolumn{2}{c}{\it{Single Dataset Training}}\\
		\hline
		TransVG~\cite{transvg} & 180 \\
		SeqTR~\cite{seqtr} & 60 \\
		Dynamic MDETR~\cite{dynamicmdetr} & 90 \\
		EEVG~\cite{eevg} & 150 \\
		\hline
		\multicolumn{2}{c}{\it{Mixed Dataset with Pre-training and Fine-tuning}}\\
		\hline
		PolyFormer~\cite{polyformer} & 20+100 \\
		OneRef~\cite{oneref} & 110+20 \\
		$\text{C}^3\text{VG}$ & 30+0 \\
		\noalign{\hrule height 1.5pt}
	\end{tabular}
	\vspace{-5pt}
	\caption{Comparison of the proposed $C^3VG$ with existing methods in terms of the number of epochs required for training.}
	\vspace{-10pt}
	\label{tab:train_epoch}
\end{table}


\section{Limitation}
When the model encounters ambiguity, its predictions tend to target a location that does not correspond to any specific object between two potential targets. This issue may arise from optimization misdirection due to the loss function's construction during model training. Unlike general object detection tasks, which involve multiple targets and often include confidence scores, REC task involve a single target, leading to a loss of clear referentiality. When the model fails to make a decisive prediction, it opts for a middle ground, yielding a lower loss but resulting in a suboptimal and unacceptable outcome.
Given the quadratic relationship between ViT's computational complexity and input scale, we utilize a relatively small input size of 320$\times$320 to maintain inference speed. Larger input sizes would slow down inference. The primary consequence of smaller inputs is coarser and less refined outputs, which is a limitation of our approach. Nonetheless, our method's ability to achieve state-of-the-art performance with this small input size demonstrates the effectiveness of $\text{C}^3\text{VG}$.

\section{More Visualization Results}
\subsection{Prediction Visualization}
\begin{figure*}[t]
	\centering
	\includegraphics[width=0.9\linewidth]{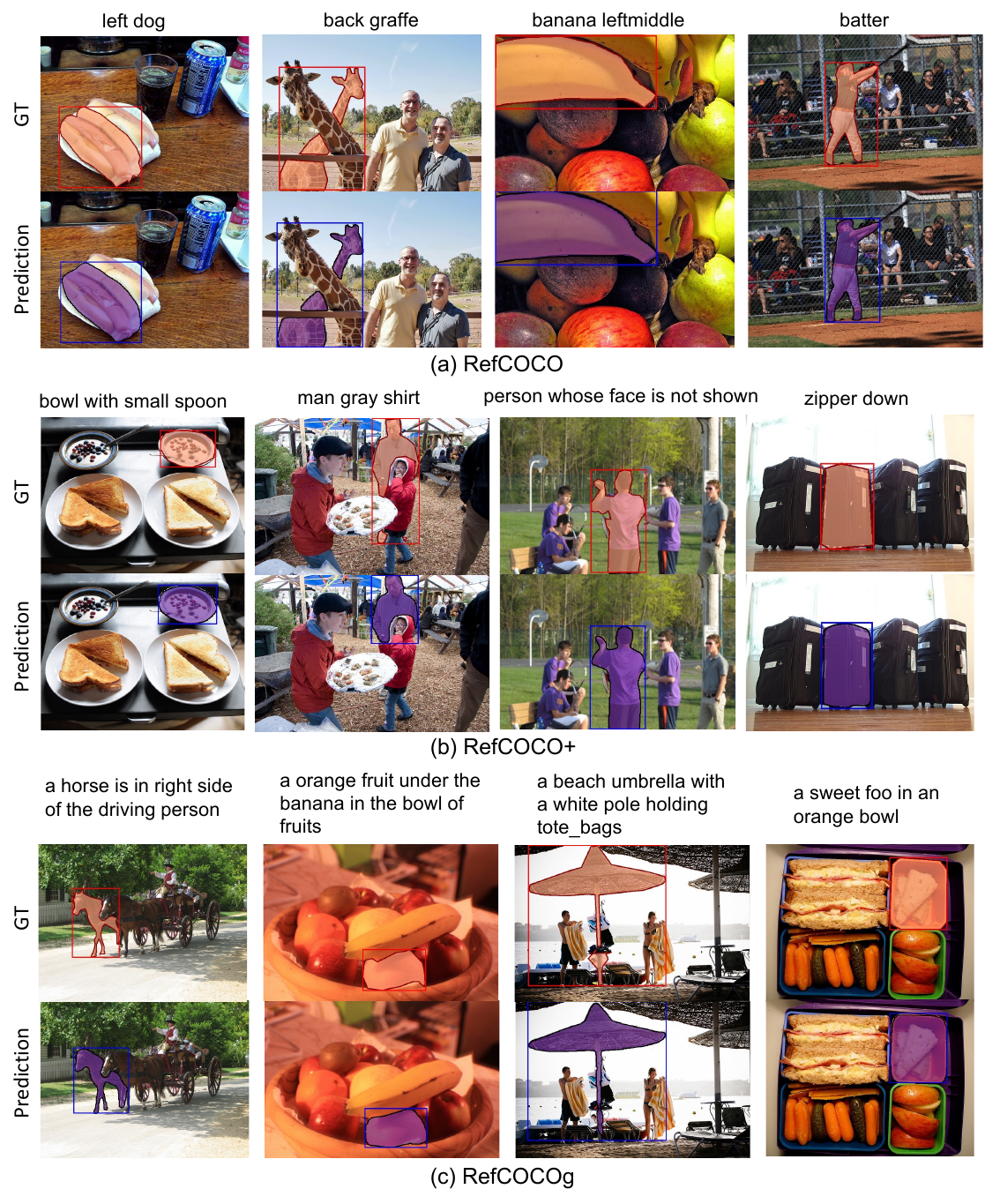} 
	\caption{Visualization of detection and segmentation results of our proposed $\text{C}^3\text{VG}$.}
	\vspace{-10pt}
	\label{fig:visualization}
\end{figure*}

Fig.~\ref{fig:visualization} visualizes the detection and segmentation results of our $\text{C}^3\text{VG}$ on the RefCOCO, RefCOCO+, and RefCOCOg datasets. Despite the increased text length and complexity in RefCOCOg, $\text{C}^3\text{VG}$ effectively leverages its powerful multimodal understanding, derived from pre-training, to handle these challenges.
\subsection{Coarse-to-Fine Visualization}
\begin{figure*}[t]
	\centering
	\includegraphics[width=0.9\linewidth]{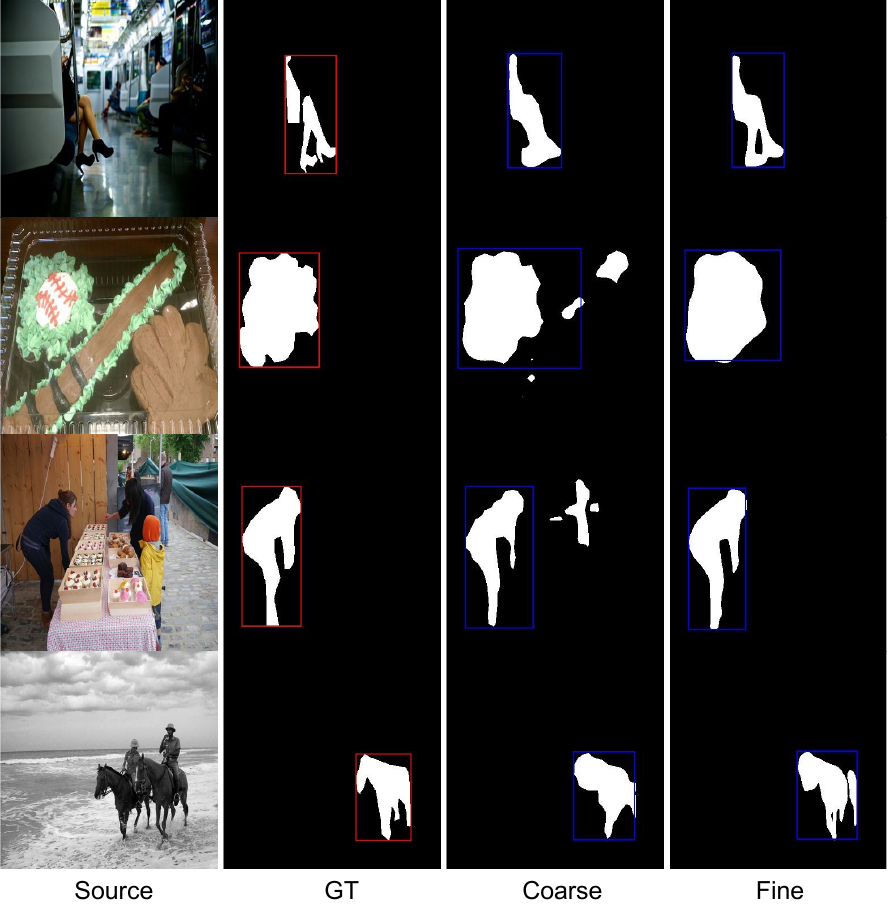} 
	\caption{Visualized detection and segmentation predictions during \textbf{early training} iterations in the RSP and RCI stages. `Coarse' represents results from the RSP stage, while `Fine' represents results from the RCI stage.}
	\vspace{-10pt}
	\label{fig:visualization_c2f}
\end{figure*}

\begin{figure*}[t]
	\centering
	\includegraphics[width=0.9\linewidth]{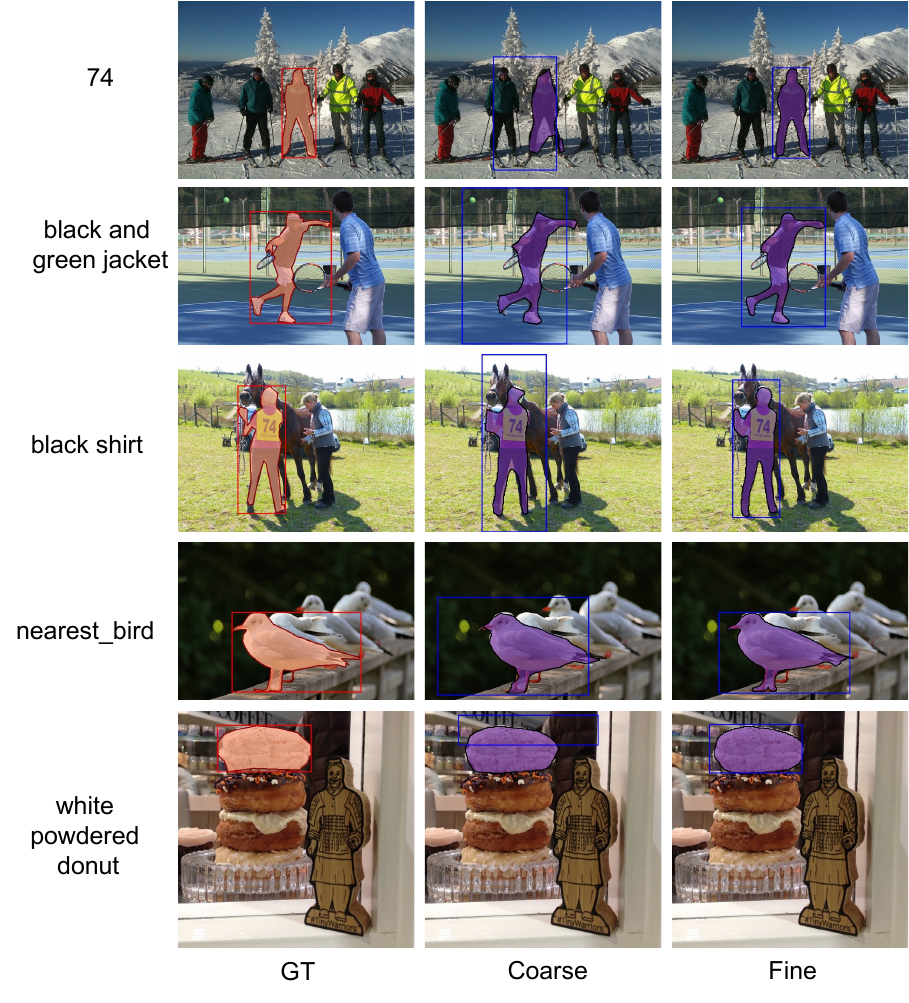} 
	\caption{Visualized detection and segmentation predictions during \textbf{testing} in the RSP and RCI stages. `Coarse' represents results from the RSP stage, while `Fine' represents results from the RCI stage.}
	\vspace{-10pt}
	\label{fig:visualization_c2f_test}
\end{figure*}

We visualize the model's output at both the early and final stages of training to observe the coarse-to-fine process in the RSP and RCI stages. Fig.~\ref{fig:visualization_c2f} shows the visualization results of the RSP and RCI stages during the early training iterations. In the RSP stage, the objective is to approximate the target's location and outline, which may lack precision. The multi-task consistency constraint subsequently refines these predictions during the RCI stage. Fig.~\ref{fig:visualization_c2f_test} illustrates the predictions of the $\text{C}^3\text{VG}$ model on the test set after training, corresponding to the RSP and RCI stages. The visualized results align with our design objectives: the RSP stage provides rough positional and semantic information, while the RCI stage offers further fine-grained localization and segmentation, incorporating consistency constraints with the coarse priors from the RSP stage.

\end{document}